\documentclass[10pt,twocolumn,letterpaper]{article}

\usepackage{iccv}
\usepackage{times}
\usepackage{comment}
\usepackage{epsfig}
\usepackage{graphicx}
\usepackage{amsmath}
\usepackage{amssymb}
\usepackage{subfigure}
\usepackage{mathrsfs}
\usepackage{algorithm,algorithmic}
\usepackage{multirow}
\usepackage{hhline}

\iccvfinalcopy 

\newcommand{\bx}{{\mathbf x}}
\newcommand{\bh}{{\mathbf h}}
\newcommand{\bz}{{\mathbf z}}

\newcommand{\bX}{{\mathbf X}}
\newcommand{\bW}{{\mathbf W}}
\newcommand{\bV}{{\mathbf V}}
\newcommand{\bTheta}{{\boldsymbol \Theta}}

\newcommand{\bbR}{{\mathbb R}}
\newcommand{\bbP}{{\mathbb P}}

\newcommand{\cL}{{\mathcal L}}
\newcommand{\cR}{{\mathcal R}}

\newcommand{\cJ}{{\mathcal J}}
\newcommand{\cQ}{{\mathcal Q}}

\newcommand{\enc}{\mathrm{enc}}
\newcommand{\dec}{\mathrm{dec}}

\renewcommand{\algorithmicrequire}{\textbf{Input:}}
\renewcommand{\algorithmicensure}{\textbf{Output:}}

\def\iccvPaperID{1294} 

\begin{document}

\title{\LaTeX\ Author Guidelines for ICCV Proceedings}

\title{Domain Generalization for Object Recognition with Multi-task Autoencoders}

\author{Muhammad Ghifary \quad W. Bastiaan Kleijn \quad Mengjie Zhang \quad David Balduzzi\\
Victoria University of Wellington\\
{\tt\small \{muhammad.ghifary,bastiaan.kleijn,mengjie.zhang\}@ecs.vuw.ac.nz, david.balduzzi@vuw.ac.nz}
}

\maketitle


\begin{abstract}
The problem of domain generalization is to take knowledge acquired from a number of related domains where training data is available, and to then successfully apply it to previously unseen domains. 
We propose a new feature learning algorithm, Multi-Task Autoencoder (MTAE), that provides good generalization performance for cross-domain object recognition.

Our algorithm extends the standard denoising autoencoder framework by substituting artificially induced corruption with naturally occurring inter-domain variability in the appearance of objects. 
Instead of reconstructing images from noisy versions, MTAE learns to transform the original image into analogs in multiple related domains. 
It thereby learns features that are robust to variations across domains. The learnt features are then used as inputs to a classifier. 

We evaluated the performance of the algorithm on benchmark image recognition datasets, where the task is to learn features from multiple datasets and to then predict the image label from unseen datasets.
We found that (denoising) MTAE outperforms alternative autoencoder-based models as well as the current state-of-the-art algorithms for domain generalization.
\end{abstract}

\vspace{-1em}
\section{Introduction}
Recent years have seen dramatic advances in object recognition by deep learning algorithms~\cite{Krizhevsky:2012aa,Ciresan:2012aa,Sutskever:2014}. Much of the increased performance derives from applying large networks to massive labeled datasets such as PASCAL VOC~\cite{pascal-voc-2007} and ImageNet~\cite{Krizhevsky:2009aa}. Unfortunately, \emph{dataset bias} -- which can include factors such as backgrounds, camera viewpoints and illumination -- often causes algorithms to generalize poorly \emph{across} datasets \cite{Torralba2011} and significantly limits their usefulness in practical applications. Developing algorithms that are \emph{invariant} to dataset bias is therefore a compelling problem.

\vspace{-1em}
\paragraph{Problem definition.}
In object recognition, the ``visual world" can be considered as decomposing into \emph{views} (e.g. perspectives or lighting conditions) corresponding to domains. For example, frontal-views and  $45^{\circ}$ rotated-views correspond to two different domains.
Alternatively, we can associate views or domains with standard image datasets such as PASCAL VOC2007~\cite{pascal-voc-2007}, and Office~\cite{Saenko:2010aa}.

The problem of learning from multiple \emph{source} domains and testing on unseen \emph{target} domains is referred to as \emph{domain generalization}~\cite{Blanchard2011,Muandet2013}. 
A \emph{domain} is a probability distribution $\bbP_k$ from which samples $\{ x_i, y_i\}_{i=1}^{N_k}$ are drawn.
Source domains provide training samples, whereas distinct target domains are used for testing. In the standard supervised learning framework, it is assumed that the source and target domains coincide. Dataset bias becomes a significant problem when training and test domains differ: applying a classifier trained on one dataset to images sampled from another typically results in poor performance~\cite{Torralba2011,Gong:2013aa}.
The goal of this paper is to learn features that improve generalization performance across domains.

\vspace{-1em}
\paragraph{Contribution.} 
The challenge is to build a system that recognizes objects in previously \emph{unseen} datasets, given one or multiple training datasets.  We introduce Multi-task Autoencoder (MTAE), a feature learning algorithm that uses a multi-task strategy~\cite{Caruana1997,Thrun1996} to learn unbiased object features, where the \emph{task} is the data reconstruction.

Autoencoders were introduced to address the problem of `backpropagation without a teacher' by using \emph{inputs as labels} -- and learning to reconstruct them with minimal distortion \cite{rumelhart:86a,Bengio:2007aa}. Denoising autoencoders in particular are a powerful basic circuit for unsupervised representation learning \cite{Vincent:2008aa}. Intuitively, corrupting inputs forces autoencoders to learn representations that are robust to noise. 

This paper proposes a broader view: that autoencoders are \emph{generic circuits for learning invariant features}. The main contribution is a new training strategy based on naturally occurring transformations such as: rotations in viewing angle, dilations in apparent object size, and shifts in lighting conditions. The resulting Multi-Task Autoencoder learns features that are robust to real-world image variability, and therefore generalize well across domains.
Extensive experiments show that MTAE with a denoising criterion outperforms the prior state-of-the-art in domain generalization over various cross-dataset recognition tasks.

\section{Related work}
Domain generalization has recently attracted attention in classification tasks, including automatic gating of flow cytometry data~\cite{Blanchard2011,Muandet2013} and object recognition~\cite{Fang2013,Khosla2012,Xu2014}.
Khosla \etal~\cite{Khosla2012} proposed a multi-task max-margin classifier, which we refer to as Undo-Bias, that explicitly encodes dataset-specific biases in feature space.
These biases are used to push the dataset-specific weights to be similar to the global weights.
Fang \etal~\cite{Fang2013} developed Unbiased Metric Learning (UML) based on learning to rank framework. 
Validated on weakly-labeled web images, UML produces a less biased distance metric that provides good object recognition performance.
and validated on weakly-labeled web images.
More recently, Xu \etal~\cite{Xu2014} extended an exemplar-SVM to domain generalization by adding a nuclear norm-based regularizer that captures the likelihoods of all positive samples.
The proposed model is denoted by LRE-SVM.

Other works in object recognition exist that address a similar problem, in the sense of having unknown targets, where the unseen dataset contains noisy images that are not in the training set~\cite{Ghifary2014,Tang:2010aa}.
However, these were designed to be noise-specific and may suffer from dataset bias when observing objects with different types of noise.

A closely related task to domain generalization is domain adaptation, where unlabeled samples from the target dataset are available during training.
Many domain adaptation algorithms have been proposed for object recognition (see, \eg, \cite{Baktashmotlagh:2014aa,Saenko:2010aa}).
Domain adaptation algorithms are not readily applicable to domain generalization, since no information is available about the target domain.

Our proposed algorithm is based on the feature learning approach.
Feature learning has been of a great interest in the machine learning community since the emergence of deep learning (see \cite{Bengio2013} and references therein).
Some feature learning methods have been successfully applied to domain adaptation or transfer learning applications~\cite{Chen:2012ab,Donahue:2014aa}.
To our best knowledge, there is no prior work along these lines on the more difficult problem of domain generalization, \ie, to create useful representations without observing the target domain.

\section{The Proposed Approach}
\label{sec:mtae}
Our goal is to learn features that provide good domain generalization.
To do so, we extend the autoencoder~\cite{Bourlard:1988aa} into a model that jointly learns multiple data-reconstruction tasks taken from related domains.
Our strategy is motivated by prior work demonstrating that learning from multiple related tasks can improve performance on a novel, yet related, task -- relative to methods trained on a single-task~\cite{Argyriou2008,Baxter2000,Caruana1997,Thrun1996}.

\subsection{Autoencoders}
Autoencoders (AE) have become established as a pretraining model for deep learning~\cite{Bengio:2007aa}.
The autoencoder training consists of two stages: 1) \emph{encoding} and 2) \emph{decoding}.
Given an unlabeled input $\bx \in \bbR^{d_x}$, a single hidden layer autoencoder $f_{\bTheta}(\bx) : \bbR^{d_x} \rightarrow \bbR^{d_x}$ can be formulated as
\begin{eqnarray}
	\label{eq:ae_rec}
	\bh &=& \sigma_{\enc}(\bW^{\top} \bx) \nonumber \\
	\hat{\bx} &=& \sigma_{\mathrm{\dec}}(\bV^{\top} \bh) = f_{\bTheta}(\bx),
\end{eqnarray}
where $\bW \in \bbR^{d_x \times d_y}$,  $\bV \in \bbR^{d_y \times d_x}$ are \emph{input-to-hidden} and \emph{hidden-to-output} connection weights\footnote{While the bias terms are incorporated in our experiments, they are intentionally omitted from equations for the sake of simplicity. } respectively,
$\bh \in \bbR^{d_h}$ is the hidden node vector,
and $\sigma_{\enc}(\cdot) = [s_{\enc}(z_1), ..., s_{\enc}(z_{d_h})]^{\top}, \sigma_{\dec}(\cdot) =  [s_{\dec}(z_1), ..., s_{\dec}(z_{d_x})]^{\top} $ are element-wise non-linear activation functions, and $s_{\enc}$ and $s_{\dec}$ are not necessarily identical.
Popular choices for the activation function $s(\cdot)$ are, \eg, the sigmoid $s(a) = (1+\exp(-a))^{-1}$ and the rectified linear (ReLU) $s(a) = \max(0, a)$.

Let $\bTheta = \{\bW, \bV\}$ be the autoencoder parameters and $\{ \bx_i\}_{i=1}^{N}$ be a set of $N$ input data. Learning corresponds to minimizing the following objective 
\begin{eqnarray}
	\label{eq:ae_obj}
	\hat{\bTheta} := \arg \min_{\bTheta} \sum_{i=1}^{N} \cL \left(f_{\bTheta}(\bx_{i}), \bx_{i} \right) 
	+ \eta \cR \left(\bTheta \right),
\end{eqnarray}
where $\cL(\cdot, \cdot)$ is the loss function, usually in the form of \emph{least square} or \emph{cross-entropy} loss, 
and $\cR(\cdot)$ is a regularization term used to avoid overfitting.
The objective (\ref{eq:ae_obj}) can be optimized by the backpropagation algorithm~\cite{Rumelhart:1986aa}.
If we apply autoencoders to raw pixels of visual object images, the weights $\bW$ usually form visually meaningful ``filters" that can be interpreted qualitatively.

To create a discriminative model using the learnt autoencoder model, either of the following options can be considered:
1) the feature map $\phi(\bx):= \sigma_\textrm{enc}(\hat{\bW}^\top \bx)$ is extracted and used as an input to supervised learning algorithms while keeping the weight matrix $\hat{\bW}$ fixed;
2) the learnt weight matrix $\hat{\bW}$ is used to initialize a neural network model and is updated during the supervised neural network training (\emph{fine-tuning}).

Recently, several variants such as denoising autoencoders (DAE)~\cite{Vincent:2010aa} and contractive autoencoders (CAE)~\cite{Rifai2011} have been proposed to extract features that are more robust to small changes of the input.
In DAEs, the objective is to reconstruct a \emph{clean} input $\bx$ given its \emph{corrupted} counterpart $\tilde{\bx} \sim \cQ(\tilde{\bx} | \bx)$.
Commonly used types of corruption are zero-masking, Gaussian, and salt-and-pepper noise.
Features extracted by DAE have been proven to be more discriminative than ones extracted by AE~\cite{Vincent:2010aa}.

\subsection{Multi-task Autoencoders}
\begin{figure}
	\centering
	\includegraphics[width=3in,height=1.7in]{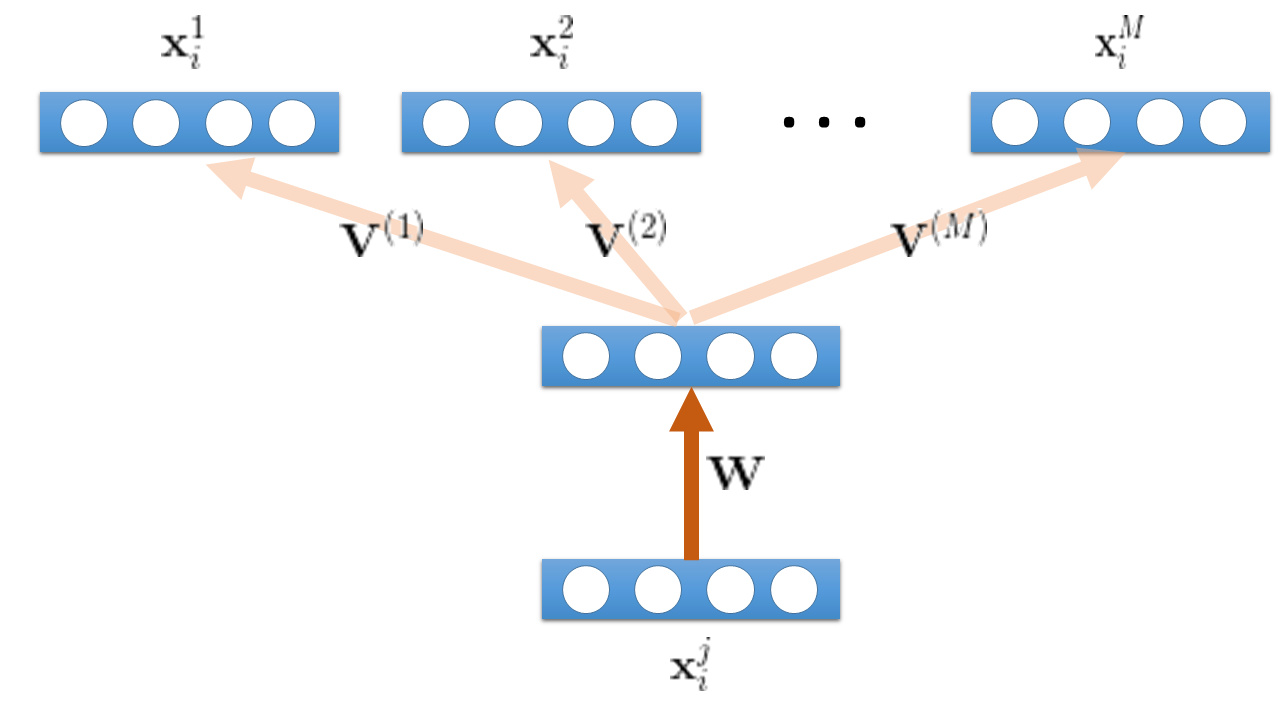}
	\caption{The Multi-task Autoencoder (MTAE) architecture, which consists of three layers with multiple separated outputs; each output corresponds to one task/domain.
	}
	\label{fig:mtae}
\end{figure}

We refer to our proposed domain generalization algorithm as \emph{Multi-task Autoencoder} (MTAE).
From an architectural viewpoint, MTAE is an autoencoder with multiple output layers, see Fig.~\ref{fig:mtae}. The input-hidden weights represent \emph{shared} parameters and the hidden-output weights represent \emph{domain-specific} parameters.
The architecture is similar to the supervised multi-task neural networks proposed by Caruana~\cite{Caruana1997}. 
The main difference is that the output layers of MTAE correspond to different domains instead of different class labels.

The most important component of MTAE is the training strategy, which constructs a generalized denoising autoencoder that learns invariances to naturally occurring transformations. Denoising autoencoders focus on the special case where the transformation is simply noise. In contrast, MTAE training treats a specific perspective on an object as the ``corrupted'' counterpart of another perspective (\eg, a rotated digit 6 is the noisy pair of the original digit).
The autoencoder objective is then reformulated along the lines of multi-task learning: the model aims to \emph{jointly achieve good reconstruction of all source views given a particular view}.
For example, applying the strategy to handwritten digit images with several views, MTAE learns representations that are invariant across the source views, see Section~\ref{sec:experiments}.

Two types of reconstruction tasks are performed during MTAE training: 
1) \textbf{\emph{self-domain}} reconstruction and 
2) \textbf{\emph{between-domain}} reconstruction.
Given $M$ source domains, there are $M\times M$ reconstruction tasks, of which $M$ task are self-domain reconstructions and the remaining $M \times (M-1)$ tasks are between-domain reconstructions.
Note that the self-domain reconstruction is identical to the standard autoencoder reconstruction (\ref{eq:ae_rec}).

\paragraph{Formal description.}
Let $\{ \bx_i^l \}_{i=1}^{n_l}$, be a set of $d_x$-dimensional data points in the $l^\text{th}$ domain, where $l \in \{1, ..., M\}$. 
Each domain's data points are combined into a matrix $\bX^l \in \bbR^{n_l \times d_x}$, where $ \bx_i^{l \top}$ is its $i^\text{th}$ row, such that $(\bx^1_i, \bx^2_i, \ldots \bx^M_i)$ form a category-level correspondence.
This configuration enforces the number of samples in a category to be the same in every domain.
Note that such a configuration is necessary to ensure that the \emph{between-domain} reconstruction works 
(we will discuss how to handle the case with unbalanced samples in Section~\ref{sec:rand_down}). 
The input and output pairs used to train MTAE can then be written as concatenated matrices 
\begin{eqnarray}
	\label{eq:rep_data}	
	\bar{\bX} &=& [\bX^1; \bX^2; ...; \bX^M], \nonumber \\
	\bar{\bX}^l &=& [\bX^l; \bX^l; ...; \bX^l]
\end{eqnarray}
where $\bar{\bX}, \bar{\bX}^l \in \bbR^{N \times d_x}$ and $N = \sum_{l=1}^{M} n_l$.
In words, $\bar{\bX}$ is the matrix of data points taken from all domains and $\bar{\bX}^l$ is the matrix of replicated data sets taken from the $l^\text{th}$ domain. 
The replication imposed in $\bar{\bX}^l$ constructs input-output pairs for the autoencoder learning algorithm. In practice, the algorithm can be implemented efficiently -- without replicating the matrix in memory.

We now describe MTAE more formally.
Let $\bar{\bx}_i^{\top}$ and $\bar{\bx}_i^{l \top}$ be the $i^\text{th}$ row of matrices $\bar{\bX}$ and $\bar{\bX}^l$, respectively, the feedforward MTAE reconstruction is
\begin{eqnarray}
\bh_{i} &=& \sigma_{\enc} (\bW^{\top} \bar{\bx}_{i}), \nonumber \\
f_{\bTheta^{(l)}}(\bar{\bx}_i) &=& \sigma_{\dec} (\bV^{(l) \top} \bh_{i}),
\label{eq:mtae_forwardpass}
\end{eqnarray}
where $\bTheta^{(l)} = \{ \bW, \bV^{(l)}\} $ contains the matrices of shared and individual weights, respectively.

The MTAE training is achieved as follows.
Let us define the loss function summed over the datapoints
\vspace{-0.5em}
\begin{eqnarray}
	J(\bTheta^{(l)}) = \sum_{i=1}^{N} \cL \left(f_{\bTheta^{(l)}}(\bar{\bx}_i ), \bar{\bx}_i^l \right) .
\end{eqnarray}
Given $M$ domains, training MTAE corresponds to minimizing the objective
\vspace{-0.5em}
\begin{eqnarray}
\label{eq:mtae_obj}
\hat{\bTheta}^{(l)} := \arg \min_{ \bTheta^{(l)} } \sum_{l=1}^{M} J(\bTheta^{(l)})+ \eta \cR(\mathbf{\Theta}^{(l)}),
\end{eqnarray}
where $\cR(\mathbf{\Theta}^{(l)})$ is a regularization term.
In this work, we use the standard $l_2$-norm weight penalty $\cR(\bTheta^{(l)}) = \| \bW \|^2_2 + \| \bV^{(l)} \|^2_2$.
Stochastic gradient descent is applied on each reconstruction task to achieve the objective (\ref{eq:mtae_obj}).
Once training is completed, 
the optimal \emph{shared} weights $\hat{\bW}$ are obtained.
The stopping criterion is empirically determined by monitoring the average loss over all reconstruction tasks during training -- 
the process is stopped when the average loss stabilizes.
The detailed steps of MTAE training is summarized in Algorithm~\ref{alg:mtae}.
\begin{algorithm}[!htb]
\footnotesize
	 \caption{The MTAE feature learning algorithm.} 
	\label{alg:mtae}
	\algorithmicrequire\; \\
	$\bullet$ Data matrices based on (\ref{eq:rep_data}): $\bar{\bX}$ and $\bar{\bX}^l, \forall l \in \{ 1,..., M\}$; \\
	$\bullet$ Source labels: $\{ y_i^l \}_{i=1}^{n_l}, \forall l \in \{1,...,M \}$; \\
	$\bullet$ The learning rate: $\alpha$;
	\begin{algorithmic}[1]
		\STATE Initialize $\bW \in \bbR^{d_x \times d_h}$ and $\bV^{(l)} \in \bbR^{d_h \times d_x}$, $\forall l\in\{1,...,M\}$ with small random real values;\\
		\WHILE{not end of epoch} 
			\STATE{Do \textsc{rand-sel} as described in Section~\ref{sec:rand_down} to balance the number of samples per categories in $\bar{\bX}$ and $\bar{\bX}^l$};
			\FOR{$l=1$ \TO $M$} 
				\FORALL{row of $\tilde{\bX}$}
					\STATE{Do a forward pass based on (\ref{eq:mtae_forwardpass});}
					\STATE{Update $\bW$ and $\bV^{(l)}$ to achieve the objective (\ref{eq:mtae_obj}) with respect to the following rules}
					{\small
					\begin{eqnarray}
						V^{(l)}_{ij} &\leftarrow& V^{(l)}_{ij} - \alpha \frac{\partial J( \{ \bW, \bV^{(l)} \})}{\partial V^{(l)}_{ij}}, \nonumber \\ 
						W_{ij} &\leftarrow& W_{ij} - \alpha \frac{\partial J(\{ \bW, \bV^{(l)} \})}{\partial W_{ij}}; \nonumber 
					\end{eqnarray}
					}
				\ENDFOR
			\ENDFOR
		\ENDWHILE
	\end{algorithmic}
\algorithmicensure\\
$\bullet$ MTAE learnt weights: $\hat{\bW}$ $\forall l \in \{ 1,..., M\}$;
\end{algorithm}

The training protocol can be supplemented with a denoising criterion as in~\cite{Vincent:2010aa} to induce more robust-to-noise features.
To do so, simply replace $\bar{\bx}_{i}$ in (\ref{eq:mtae_forwardpass}) with its corrupted counterpart  $\tilde{\bar{\bx}}_i \sim Q(\tilde{\bar{\bx}}_i | \bar{\bx}_i)$.
We name the MTAE model after applying the denoising criterion the \emph{Denoising Multi-task Autoencoder} (D-MTAE).

\subsection{Handling unbalanced samples per category}
\label{sec:rand_down}
%
%

MTAE requires that every instance in a particular domain has a category-level corresponding pair in every other domain. MTAE's apparent applicability is therefore limited to situations where the number of source samples per category is the same in every domain. 
However, unbalanced samples per category occur frequently in applications. 
To overcome this issue, we propose a simple \emph{random selection} procedure applied in the \emph{between-domain} reconstructions, denoted by \textsc{rand-sel}, which is simply balancing the samples per category while keeping their category-level correspondence.

In detail, the \textsc{rand-sel} strategy is as follows. 
Let $m_c$ be the number of subsamples in the $c$-th category, where $m_c = \min(n_{1c},n_{2c},\ldots,n_{Mc})$ and $n_{lc}$ is the number of samples in the $c$-th category of domain $l \in \{1,\ldots,M\}$.
For each category $c$ and each domain $l$, select $m_c$ samples randomly such that $n_{lc} = n_{2c} = \ldots n_{Mc} = m_c$.
This procedure is executed in every iteration of the MTAE algorithm, see Line 3 of Algorithm~\ref{alg:mtae}.


\section{Experiments and Results}
\label{sec:experiments}
We conducted experiments on several real world object datasets to evaluate the domain generalization ability of our proposed system.
In Section~\ref{sec:exp1}, we investigate the behaviour of MTAE in comparison to standard single-task autoencoder models on raw pixels as proof-of-principle.
In Section~\ref{sec:exp2}, we evaluate the performance of MTAE against several state-of-the-art algorithms on modern object datasets such as the Office~\cite{Saenko:2010aa}, Caltech~\cite{Caltech}, PASCAL VOC2007~\cite{pascal-voc-2007}, LabelMe~\cite{LabelMe}, and SUN09~\cite{SUN09}.

\subsection{Cross-recognition on the MNIST and ETH-80 datasets}
\label{sec:exp1}
In this part, we aim to understand MTAE's behavior when learning from multiple domains that form physically reasonable object transformations such as roll, pitch rotation, and dilation.
The task is to categorize objects in views (domains) that were not presented during training. We evaluate MTAE against several autoencoder models.
To perform the evaluation, a variety of object views were constructed from the MNIST handwritten digit~\cite{LeCun:1998aa} and ETH-80 object~\cite{Leibe2003} datasets.

\vspace{-1em}
\paragraph{Data setup.}
\label{sec:setting}
We created four new datasets from MNIST and ETH-80 images: 1) MNIST-r, 2) MNIST-s, 3) ETH80-p, and 4) ETH80-y.
These new sets contain multiple domains so that every instance in one domain has a pair in another domain.
The detailed setting for each dataset is as follows.

\textbf{MNIST-r} contains six domains, each corresponding to a degree of roll rotation.
We randomly chose 1000 digit images of ten classes from the original MNIST training set to represent the \emph{basic} view, i.e., 0 degree of rotation;\footnote{Note that the rotation angle of the basic view is not perfectly $0^{\circ}$ since the original MNIST images have varying appearances.}
each class has 100 images.
Each image was subsampled to a $16\times16$ representation to simplify the computation.
This subset of 1000 images is denoted by $M$.
We then created 5 rotated views from $M$ with $15^{\circ}$ difference in counterclockwise direction, denoted by $M_{15^{\circ}}$, $M_{30^{\circ}}$. $M_{45^{\circ}}$, $M_{60^{\circ}}$, and $M_{75^{\circ}}$.
The \textbf{MNIST-s} is the counterpart of MNIST-r, where each domain corresponds to a dilation factor.
The views are denoted by $M$, $M_{*0.9}$, $M_{*0.8}$, $M_{*0.7}$, and $M_{*0.6}$, where the subscripts represent the dilation factors with respect to $M$.

\begin{figure}
	\centering
	\begin{center}
		\subfigure[$M$]{\includegraphics[width=0.48in]{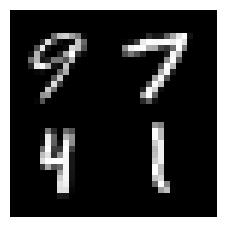}}  
		\subfigure[$M_{15^{\circ}}$]{\includegraphics[width=0.48in]{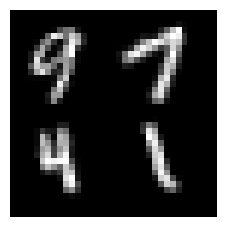}}
		\subfigure[$M_{30^{\circ}}$]{\includegraphics[width=0.48in]{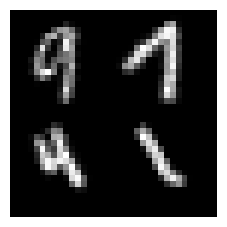}}
		\subfigure[$M_{45^{\circ}}$]{\includegraphics[width=0.48in]{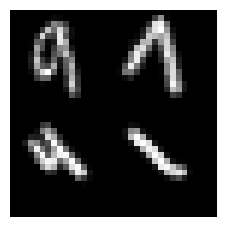}}
		\subfigure[$M_{60^{\circ}}$]{\includegraphics[width=0.48in]{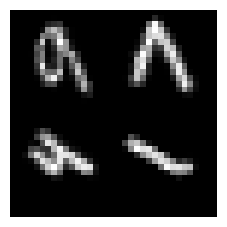}}
		\subfigure[$M_{75^{\circ}}$]{\includegraphics[width=0.48in]{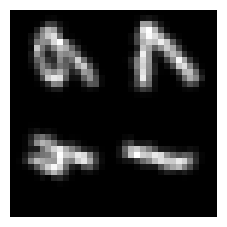}}
		\subfigure[$M$]{\includegraphics[width=0.56in]{Xm0_samples.png}} 
		\subfigure[$M_{*0.9}$]{\includegraphics[width=0.54in]{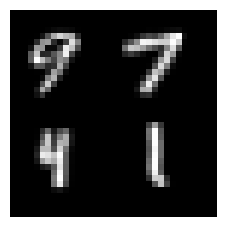}}
		\subfigure[$M_{*0.8}$]{\includegraphics[width=0.54in]{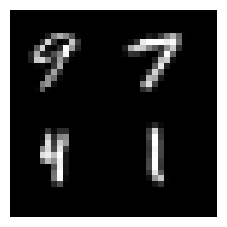}}
		\subfigure[$M_{*0.7}$]{\includegraphics[width=0.54in]{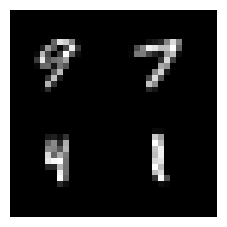}}
		\subfigure[$M_{*0.6}$]{\includegraphics[width=0.54in]{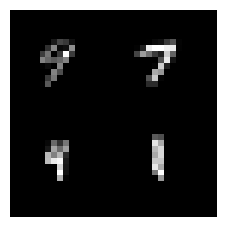}}
		\subfigure[$E_{p0^{\circ}}$]{\includegraphics[width=0.54in]{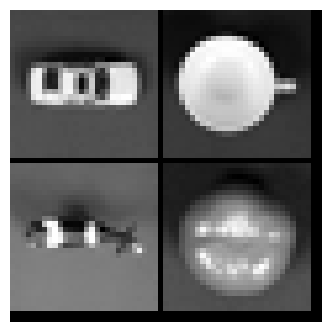}}  
		\subfigure[$E_{p22^{\circ}}$]{\includegraphics[width=0.54in]{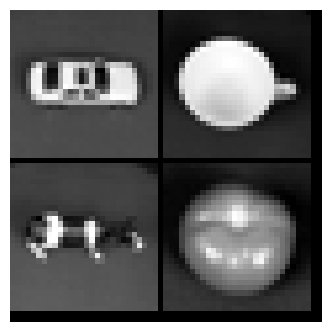}}
		\subfigure[$E_{p45^{\circ}}$]{\includegraphics[width=0.54in]{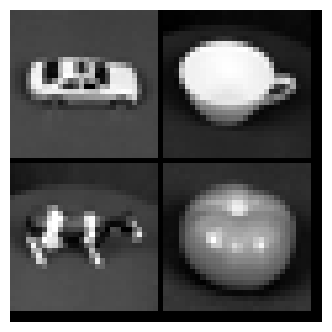}}
		\subfigure[$E_{p68^{\circ}}$]{\includegraphics[width=0.54in]{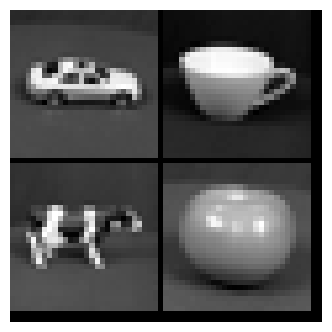}}
		\subfigure[$E_{p90^{\circ}}$]{\includegraphics[width=0.54in]{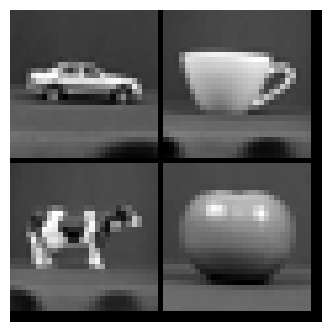}}
	\end{center}
	\vspace{-1em}
	\caption{Some image examples from the MNIST-r, MNIST-s, and ETH80-p .}
	\label{fig:rotMNIST}
\end{figure}

The \textbf{ETH80-p} consists of eight object classes with 10 subcategories for each class.
In each subcategory, there are 41 different views with respect to pose angles.
We took five views from each class denoted by $E_{p0^{\circ}}$, $E_{p22^{\circ}}$, $E_{p45^{\circ}}$, $E_{p68^{\circ}}$, and $E_{p90^{\circ}}$, which represent the horizontal poses, i.e., pitch-rotated views starting from the top view to the side view.
This makes the number of instances only 80 for each view.
We then greyscaled and subsampled the images to $28\times28$.
The \textbf{ETH80-y} contains five views of the ETH-80 representing the vertical poses, i.e., yaw-rotated views starting from the right-side view to the left-side view denoted by $E_{+y90^{\circ}}$, $E_{+y45^{\circ}}$, $E_{y0^{\circ}}$, $E_{-y45^{\circ}}$, and $E_{-y90^{\circ}}$.
Other settings such as the image dimensionality and preprocessing stage are similar to ETH80-p.
Examples of the resulting views are depicted in Fig.~\ref{fig:rotMNIST}.

\begin{table*}[!htb]
\centering
	\caption{The \emph{leave-one-\textbf{domain}-out} classification accuracies \% on the MNIST-r and MNIST-s. 
	Bold-red and bold-black indicate the best and second best performance.
}
	\scalebox{0.8}{
	\begin{tabular}{ c | c || c | c | c | c | c | c | c }
		\hline
		Source & Target & Raw &  AE & DAE & CAE & uDICA & \textbf{MTAE} & \textbf{D-MTAE}\\
    		\hline
    		\multicolumn{9}{c}{MNIST-r leave-one-\textbf{roll-rotation}-out} \\
    		\hline
		$M_{15^{\circ}}$, $M_{30^{\circ}}$, $M_{45^{\circ}}$, $M_{60^{\circ}}$, $M_{75^{\circ}}$ & $M$ &
		  $52.40$  &  $74.20$ & $76.90$ & $72.10$ & $67.20$ & \underline{$\mathbf{77.90}$} & {\color{red} \underline{$\mathbf{82.50}$}}\\
		$M$, $M_{30^{\circ}}$, $M_{45^{\circ}}$, $M_{60^{\circ}}$, $M_{75^{\circ}}$ & $M_{15^{\circ}}$ &
		  $74.10$  &  $93.20$ & $93.20$ & $95.30$ & $87.80$ & \underline{$\mathbf{95.70}$} &  {\color{red} \underline{$\mathbf{96.30}$}}\\
		$M$, $M_{15^{\circ}}$, $M_{45^{\circ}}$, $M_{60^{\circ}}$, $M_{75^{\circ}}$ & $M_{30^{\circ}}$ &
		  $71.40$  &  $89.90$ & $91.30$ & \underline{$\mathbf{92.60}$} & $88.80$ & $91.20$ &  {\color{red} \underline{$\mathbf{93.40}$}}\\
		$M$, $M_{15^{\circ}}$, $M_{30^{\circ}}$, $M_{60^{\circ}}$, $M_{75^{\circ}}$ & $M_{45^{\circ}}$ &
		  $61.40$  &  {\color{red}\underline{$\mathbf{82.20}$}} & $81.10$ & \underline{$\mathbf{81.50}$} & $77.80$ & $77.30$ &  $78.60$ \\
		$M$, $M_{15^{\circ}}$, $M_{30^{\circ}}$, $M_{45^{\circ}}$, $M_{75^{\circ}}$ & $M_{60^{\circ}}$ &
		  $67.40$ &  $90.00$ & \underline{$\mathbf{92.80}$} & $92.70$ &$84.20$ & $92.40$ &  {\color{red} \underline{$\mathbf{94.20 }$}}\\
		$M$, $M_{15^{\circ}}$, $M_{30^{\circ}}$, $M_{45^{\circ}}$, $M_{60^{\circ}}$ & $M_{75^{\circ}}$ &
		  $55.40$ &  $73.80$ & $76.50$ & $79.30$ & $69.50$ & \underline{$\mathbf{79.90}$}&   {\color{red}\underline{$\mathbf{80.50}$}} \\
		\hline
		\hline
		\multicolumn{2}{c ||}{Average} & 
		  $63.68$ & $83.88$ & $85.30$ & $85.58$ & $79.22$ & \underline{$\mathbf{85.73}$} & {\color{red} \underline{$\mathbf{87.58}$}} \\
		\hline
		\hline
		\multicolumn{9}{c}{MNIST-s leave-one-\textbf{dilation}-out} \\
		\hline
		$M_{*0.9}$, $M_{*0.8}$, $M_{*0.7}$, $M_{*0.6}$ & $M$ &
		  $54.00$  &  $67.50$ & $71.80$ & \underline{$\mathbf{75.80}$} &  \underline{$\mathbf{75.80}$} & $74.50$ & {\color{red} \underline{$\mathbf{76.00}$}}\\
		$M$, $M_{*0.8}$, $M_{*0.7}$, $M_{*0.6}$ & $M_{*0.9}$ &
		  $80.40$ &  $95.10$ & $94.00$ & $94.90$ & $88.60$ & \underline{$\mathbf{97.80}$} &  {\color{red} \underline{$\mathbf{98.00}$}} \\
		$M$, $M_{*0.9}$, $M_{*0.7}$, $M_{*0.6}$ & $M_{*0.8}$ &
		  $82.60$ &  $94.60$ & $92.90$ &  $94.90$ & $86.60$ &\underline{$\mathbf{96.30}$} &  {\color{red}\underline{$\mathbf{96.40}$}} \\
		$M$, $M_{*0.9}$, $M_{*0.8}$, $M_{*0.6}$ & $M_{*0.7}$ &
		  $78.20$ & $93.70$ & $91.60$ & $92.50$ & $87.40$ & {\color{red}\underline{$\mathbf{95.80}$}} &  \underline{$\mathbf{94.90}$} \\
		$M$, $M_{*0.9}$, $M_{*0.8}$, $M_{*0.7}$ & $M_{*0.6}$ &
		  $64.70$ & $74.80$ & $76.10$ & $77.50$ & $75.30$ &  \underline{$\mathbf{78.00}$} &  {\color{red}\underline{$\mathbf{78.30}$}} \\
		\hline
		\hline
		\multicolumn{2}{c ||}{Average} & 
		  $71.98$ & $85.14$ & $85.28$ &  $87.12$ & $82.74$ & \underline{$\mathbf{88.48}$} & {\color{red} \underline{$\mathbf{88.72}$}} \\	
		\hline
	\end{tabular}
	}
	\label{tab:mnist_rot_acc}
\end{table*}

\vspace{-1em}
\paragraph{Baselines.}
We compared the classification performance of our models with several single-task autoencoder models:
Descriptions of the methods and their hyperparameter settings are provided below.
\begin{itemize}
\itemsep0em
\item \textbf{AE}~\cite{Bengio:2007aa}: 
	the standard \emph{autoencoder} model trained by stochastic gradient descent, where all object views were concatenated as one set of inputs.
	The number of hidden nodes was fixed at 500 on the MNIST dataset and at 1000 on the ETH-80 dataset.
	The learning rate, weight decay penalty, and number of iterations were empirically determined at $0.1$, $3\times10^{-4}$, and $100$, respectively.
\item \textbf{DAE}~\cite{Vincent:2010aa}: 
	the \emph{denoising autoencoder} with zero-masking noise, where all object views were concatenated as one set of input data.
	The corruption level was fixed at $30\%$ for all cases.
	Other hyper-parameter values were identical to AE.
\item \textbf{CAE}~\cite{Rifai2011}:
	the autoencoder model with the Jacobian matrix norm regularization referred to as the \emph{contractive autoencoder}.
	The corresponding regularization constant $\lambda$ was set at 0.1.
\item \textbf{MTAE}: 
	our proposed multi-task autoencoder model with identical hyper-parameter settings as AE, except for the learning rate set at 0.03, which was also chosen empirically. 
	This value provides a lower reconstruction error for each task and visually clearer first layer weights.
\item \textbf{D-MTAE}: MTAE with a denoising criterion.
	The learning rate was set the same as MTAE; other hyper-parameters followed DAE.
\end{itemize}

We also evaluated the unsupervised Domain-Invariant Component Analysis (\textbf{uDICA})~\cite{Muandet2013} on these datasets for completness.
The hyper-parameters were tuned using 10-fold cross-validation on source domains.
We also did experiments using the supervised variant, DICA, with the same tuning strategy.
Surprisingly, the peak performance of uDICA is consistently higher than DICA. 
A possible explanation is that the Dirac kernel function measuring the label similarity is less appropriate in this application.	

We normalized the raw pixels to a range of $[0,1]$ for autoencoder-based models and $l_2$-unit ball for uDICA.
We evaluated the classification accuracies of the learnt features using multi-class SVM with linear kernel (L-SVM)~\cite{Crammer2001}.
Using a linear kernel keeps the classifier simple -- since our main focus is on the feature extraction process.
The LIBLINEAR package~\cite{Fan:2008aa} was used to run the L-SVM.

\vspace{-1em}
\paragraph{Cross-domain recognition results.}
We evaluated the object classification accuracies of each algorithm by \emph{leave-one-domain-out} test, \ie, taking one domain as the test set and the remaining domains as the training set.
For all autoencoder-based algorithms, we repeated the experiments on each \emph{leave-one-domain-out} case 30 times and reported the average accuracies.
The standard deviations are not reported since they are small ($\pm 0.1$).

The detailed results on the MNIST-r and MNIST-s can be seen in Table \ref{tab:mnist_rot_acc}.
On average, MTAE has the second best classification accuracies, and in particular outperforms single-task autoencoder models.
This indicates that the multi-task feature learning strategy can provide better discriminative features than the single-task feature learning \wrt unseen object views.

The algorithm with the best performance is on these datasets is D-MTAE.
Specifically, D-MTAE performs best on average and also on 9 out of 11 individual cross-domain cases of the MNIST-r and MNIST-s.
The closest single-task feature learning competitor to D-MTAE is CAE.
This suggests that the denoising criterion strongly benefits domain generalization.
The denoising criterion is also useful for single-task feature learning although it does not yield competitive accuracies, see AE and DAE performance.

We also obtain a consistent trend on the ETH80-p and ETH80-y datasets, \ie, D-MTAE and MTAE are the best and second best models.
In detail, D-MTAE and MTAE produce the average accuracies of $87.85\%$ and  $87.50\%$ on the ETH80-p, and $97\%$ and $96.50\%$ on the ETH80-y.

Observe that there is an anomaly in the MNIST-r dataset: the performance on $M_{45^{\circ}}$ is far worse than its neighbors ($M_{30^{\circ}}, M_{60^{\circ}}$).
This anomaly appears to be related to the geometry of the MNIST-r digits. 
We found that the most frequently misclassified digits are 4, 6, and 9 on $M_{45^{\circ}}$, which rarely occurs on other MNIST-r's domains -- typically 4 as 9, 6 as 4, and 9 as 8. The same phenomenon applies to L-SVM.

\begin{figure*}
	\centering
	\begin{center}
		\subfigure[AE]{\includegraphics[width=1.2in]{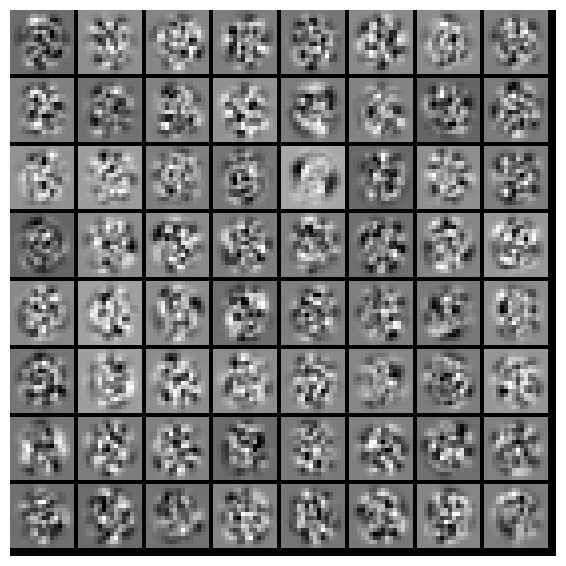}} \quad
		\subfigure[DAE]{\includegraphics[width=1.2in]{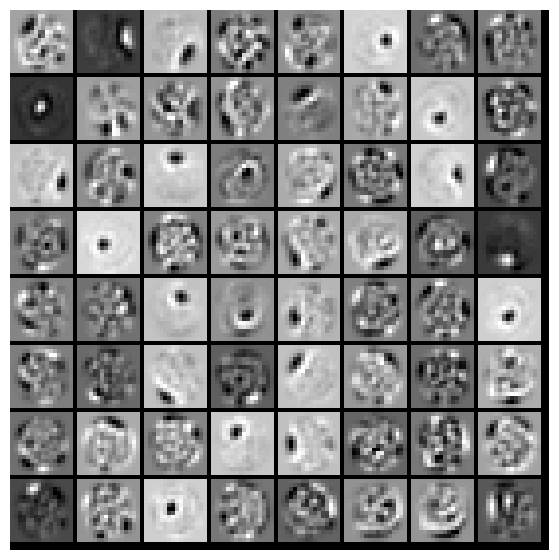}} \quad
		\subfigure[MTAE]{\includegraphics[width=1.2in]{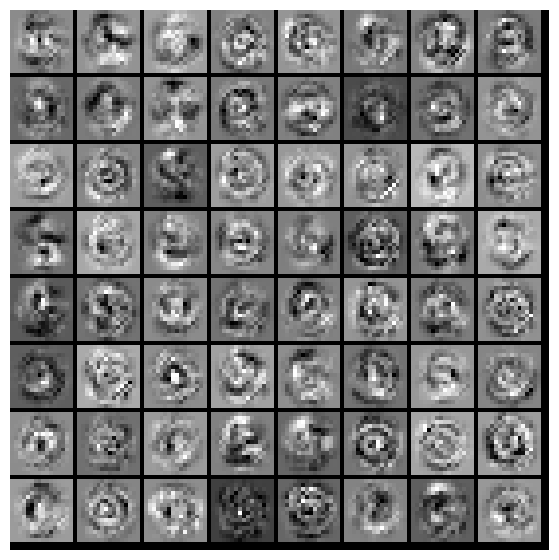}} \quad
		\subfigure[D-MTAE]{\includegraphics[width=1.2in]{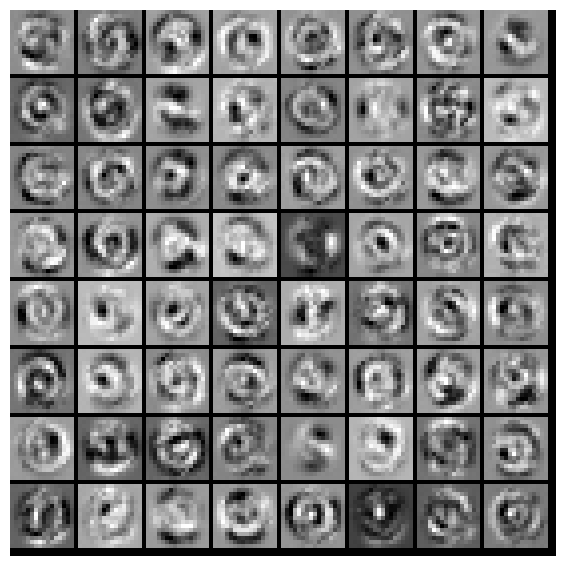}}
	\end{center}
\vspace{-1em}
	\caption{The 2D visualization of 100 randomly chosen weights after pretraining on the MNIST-r dataset. 
		  Each patch corresponds to a row of the learnt weight matrix $\mathbf{W}$ that represents a ``filter''.
		  The weight value $w_{ij} \geq 3$ is depicted with white, $w_{ij} \leq -3$ is depicted with black, 
		  otherwise it is gray.
	}
	\vspace{-0.5em}
	\label{fig:filters}
\end{figure*}

\paragraph{Weight visualization.}
Useful insight is obtained from considering the qualitative outcome of the MTAE training by visualizing the first layer weights.
Figure~\ref{fig:filters} depicts the weights of some autoencoder models, including ours, on the MNIST-r dataset.
\emph{Both MTAE and D-MTAE's weights form ``filters'' that tend to capture the underlying transformation across the MNIST-r views, which is the rotation.}
This effect is unseen in AE and DAE, the filters of which only explain the \emph{contents} of handwritten digits in the form of Fourier component-like descriptors such as local blob detectors and stroke detectors~\cite{Vincent:2010aa}.
This might be a reason that MTAE and D-MTAE features can provide better domain generalization than AE and DAE, since they implicitly capture the relationship among the source domains.

Next we discuss the difference between MTAE and D-MTAE filters.
The D-MTAE filters not only capture the object transformation, but also produce features that describe the object contents more distinctively.
These filters basically combine both properties of the DAE and MTAE filters that might benefit the domain generalization.

\vspace{-1em}
\begin{figure}[!htb]
	\centering
	\includegraphics[width=2.9in]{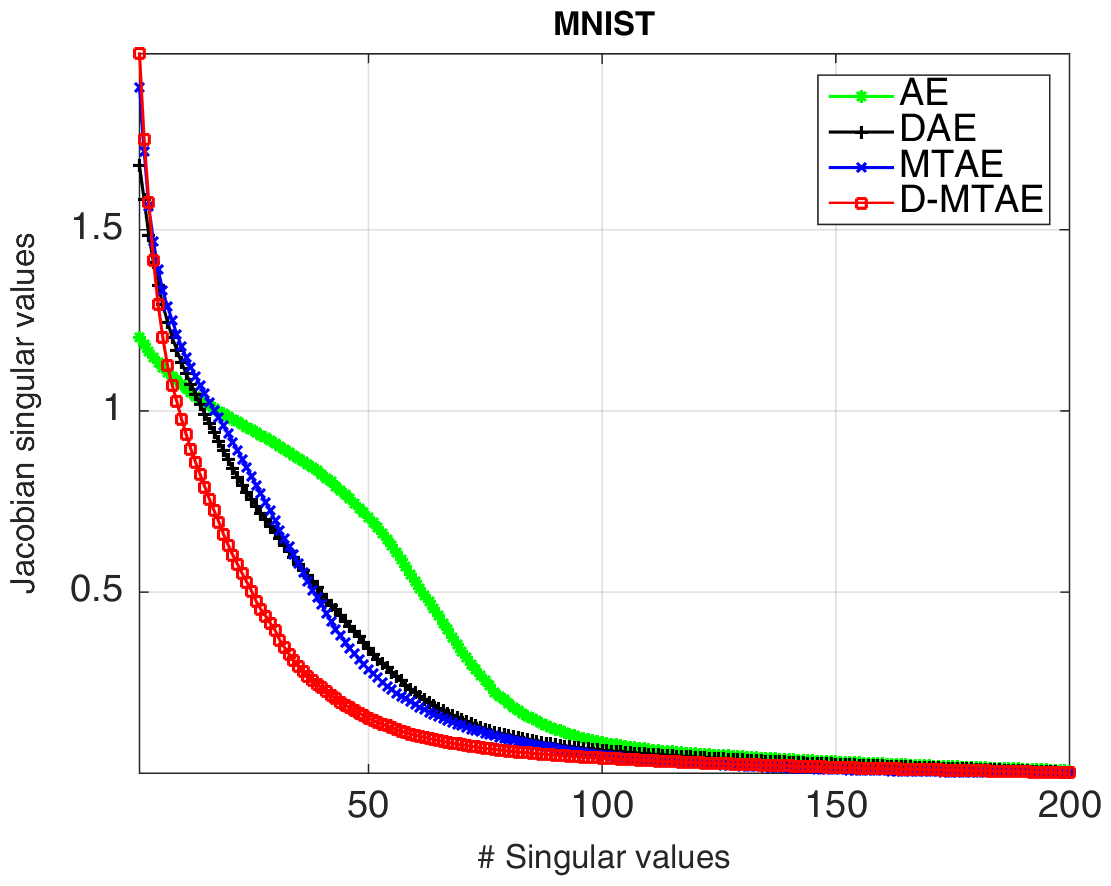}
	\caption{The average singular value spectrum of the Jacobian matrix over the MNIST-r and MNIST-s datasets.}
	\vspace{-0.5em}
	\label{fig:sv_jacobian}
\end{figure}

\paragraph{Invariance analysis.}
A possible explanation for the effectiveness of MTAE relates to the dimensionality of the manifold in feature space where samples concentrate. We hypothesize that if features concentrate near a low-dimensional submanifold, then the algorithm has found simple invariant features and will generalize well. 

To test the hypothesis, we examine the singular value spectrum of the Jacobian matrix $\cJ_{\bx}(\bz) = \left[ \frac{\partial z_i}{\partial x_j} \right]_{ij} $, where $\bx$ and $\bz$ are the input and feature vectors respectively~\cite{Rifai2011}. The spectrum describes the local dimensionality of the manifold around which samples concentrate. If the spectrum decays rapidly, then the manifold is locally of low dimension.

Figure~\ref{fig:sv_jacobian} depicts the average singular value spectrum on test samples from MNIST-r and MNIST-s. 
The spectrum of D-MTAE decays the most rapidly, followed by MTAE and then DAE (with similar rates), and AE decaying the slowest. 
The ranking of decay rates of the four algorithms matches their ranking in terms of  empirical performance in Table~\ref{tab:mnist_rot_acc}.
Figure~\ref{fig:sv_jacobian} thus provides partial confirmation for our hypothesis. However, a more detailed analysis is necessary before drawing strong conclusions.

\subsection{Cross-recognition on the Office, Caltech, VOC2007, LabelMe, and SUN09 datasets}
\label{sec:exp2}
In the second set of experiments, we evaluated the cross-recognition performance of the proposed algorithms on modern object datasets.
The aim is to show that MTAE and D-MTAE are applicable and competitive in the more general setting.
We used the Office, Caltech, PASCAL VOC2007, LabelMe, and SUN09 datasets from which we formed two cross-domain datasets.
Our general strategy is to extend the generalization of features extracted from the current best deep convolutional neural network~\cite{Krizhevsky:2012aa}.

\vspace{-1em}
\paragraph{Data Setup.}
The first cross-domain dataset consists of images from PASCAL VOC2007 (V), LabelMe (L), Caltech-101 (C), and SUN09 (S)
datasets, each of which represents one domain.
C is an object-centric dataset, while V, L, and S are scene-centric.
This dataset, which we abbreviate as \textbf{VLCS}, shares five object categories: `bird', `car', `chair', `dog', and `person'.
Each domain in the VLCS dataset was divided into a training set ($70\%$) and a test set ($30\%$) by random selection from the overall dataset.
The detailed training-test configuration for each domain is summarized in Table~\ref{tab:vlcs_conf}.
Instead of using the raw features directly, we employed the $\textnormal{DeCAF}_6$ features \cite{Donahue:2014aa} as inputs to the algorithms.
These features have dimensionality of 4,096 and are publicly available.\footnote{ ${\scriptstyle \mathrm{http://www.cs.dartmouth.edu/chenfang/proj\_page/FXR\_iccv13/index.php}}$}

The second cross-domain dataset is referred to as the \textbf{Office+Caltech}~\cite{Saenko:2010aa,Gong:2012aa} dataset that contains four domains: 
Amazon (A), Webcam (W), DSLR (D), and Caltech-256 (C), which share ten common categories.
This dataset has 8 to 151 instances per category per domain, and 2,533 instances in total.
We also used the $\textnormal{DeCAF}_6$ features extracted from this dataset, which are also publicly available.\footnote{${\scriptstyle \mathrm{http://vc.sce.ntu.edu.sg/transfer\_learning\_domain\_adaptation/}}$}

\begin{table}[!htb]
	\caption{The number of training and test instances for each domain in the VLCS dataset.}
	\centering
	\scalebox{0.85}{
	\begin{tabular}{r || c | c | c | c }
	\hline
	Domain & VOC2007 & LabelMe &Caltech-101& SUN09 \\
	\hline
	\#training & 2,363 & 1,859 & 991 & 2,297 \\
	\#test & 1,013 & 797 & 424 & 985 \\
	\hline
	\end{tabular}
	}
	\label{tab:vlcs_conf}
\end{table}
\vspace{-1em}
\begin{table}[!htb]
	\caption{The groundtruth L-SVM accuracies $\%$ on the standard training-test evaluation. 
The left-most column indicates the training set, while the upper-most row indicates the test set.}
	\centering
	\scalebox{0.85}{
	\begin{tabular}{ c || c | c | c | c}
	\hline
	Training/Test & VOC2007 & LabelMe & Caltech-101 & SUN09\\
	\hline
	VOC2007 & $\mathbf{66.34}$ & $34.50$ & $65.09$ & $52.49$ \\
	LabelMe & $44.03$ & $\mathbf{68.76}$ & $43.87$ & $41.02$ \\ 
	Caltech-101 & $52.81$ & $32.37$ & $\mathbf{95.99}$ & $39.29$ \\
	SUN09 & $52.42$ & $42.03$ & $40.33$ & $\mathbf{74.21}$ \\
	\hline
	\end{tabular}
	}
	\label{tab:vlcs_bias}
\end{table}
\begin{table*}[!htb]
	\caption{The cross-recognition accuracy $\%$ on the VLCS dataset.}
	\centering
	\scalebox{0.85}{
	\begin{tabular}{c | c || c | c | c | c | c | c | c | c}
	\hline
	Source & Target & L-SVM & 1HNN & DAE+1HNN &  Undo-Bias & UML &  LRE-SVM & \textbf{MTAE+1HNN} &\textbf{D-MTAE+1HNN} \\
	\hline
	L,C,S & V & $58.86$ & $59.10$ & \underline{$\mathbf{62.00}$} &$54.29$ &  $56.26$ & $60.58$  &  $61.09$ &  {\color{red} \underline{$\mathbf{63.90}$}} \\	
	V,C,S & L & $52.49$ & $58.20$ & $59.23$ & $58.09$ & $58.50$ & \underline{$\mathbf{59.74}$} & $59.24$ &  {\color{red} \underline{$\mathbf{60.13}$}} \\
	V,L,S & C & $77.67$ & $86.67$ & $90.24$ &$87.50$ & {\color{red} \underline{$\mathbf{91.13}$}} & $88.11$ &  \underline{$\mathbf{90.71}$} &  $89.05$\\
	V,L,C & S & $49.09$ & $57.86$ & 57.45 &$54.21$& $58.49$ & $54.88$ & \underline{$\mathbf{60.20}$} & {\color{red} \underline{$\mathbf{61.33}$}}\\
	\hline
	\hline
	\multicolumn{2}{c || }{Avg.} & $59.93$ & 65.46 & $67.23$ & $63.52$ & $65.85$ & $65.83$ & \underline{$\mathbf{67.81}$} & {\color{red} \underline{$\mathbf{68.60}$}} \\
	\hline
	\end{tabular}
	}
	\label{tab:vlcs_acc}
\end{table*}
\begin{table*}[!htb]
	\caption{The cross-recognition accuracy $\%$ on the Office+Caltech dataset.}
	\centering
	\scalebox{0.85}{
	\begin{tabular}{c | c || c | c | c | c | c | c | c | c}
	\hline
	Source & Target & L-SVM & 1HNN & DAE+1HNN& Undo-Bias & UML &  LRE-SVM & \textbf{MTAE+1HNN} &\textbf{D-MTAE+1HNN} \\
	\hline
	A,C & D,W & $82.08 $ & $83.41$ & $82.05$ & $80.49$ & $82.29$ &\underline{$\mathbf{84.59}$} & $84.23$ & {\color{red} \underline{$\mathbf{85.35}$}}\\
	D,W & A,C & $76.12$  & $76.49$ & $79.04$ & $69.98$ & $79.54$ & {\color{red} \underline{$\mathbf{81.17}$}} & $79.30$ & \underline{$\mathbf{80.52}$}\\
	C,D,W & A & $90.61$  & $92.13$ & $92.02$ & $90.98$ & $91.02$ &$91.87$ & \underline{$\mathbf{92.20}$} & {\color{red} \underline{$\mathbf{93.13}$}}\\
	A,W,D & C & $84.51$  & $85.89$ & $85.17$ & $85.95$  & $84.59$ & {\color{red} \underline{$\mathbf{86.38}$}} & $85.98$ & \underline{$ \mathbf{86.15}$} \\
	\hline
	\hline
	\multicolumn{2}{c ||}{Avg.} & $83.33$  & $84.48$ & $84.70$ & $81.85$ & $84.36$  & \underline{$\mathbf{86.00}$} & $85.43$ & {\color{red} \underline{$\mathbf{86.29}$}}\\
	\hline
	\end{tabular}
	}
	\label{tab:office_acc}
\end{table*}

\vspace{-1em}
\paragraph{Training protocol.}
On these datasets, we utilized the MTAE or D-MTAE learning as \emph{pretraining} for a fully-connected neural network with one hidden layer (1HNN).
The number of hidden nodes was set at 2,000, which is less than the input dimensionality.
In the pretraining stage, the number of output layers was the same as the number of source domains --each corresponds to a particular source domain.
The sigmoid activation and linear activation functions were used for $\sigma_{\enc}(\cdot)$ and $\sigma_{\dec}(\cdot)$.

The MTAE pretraining was run with the learning rate at $5 \times 10^{-4}$, the number of epochs at $500$, and the batch size at $10$, which were empirically determined \wrt the smallest average reconstruction loss.
D-MTAE has the same hyper-parameter setting as MTAE except the additional zero-masking corruption level at $20\%$.
After the pretraining is completed, we then performed back-propagation \emph{fine-tuning} using 1HNN with softmax output, where the first layer weights were initialized by either the MTAE or D-MTAE learnt weights.
The supervised learning hyper-parameters were tuned using 10-fold cross validation (10FCV) on source domains.
We denote the overall models by \textbf{MTAE+1HNN} and  \textbf{D-MTAE+1HNN}.

\vspace{-1em}
\paragraph{Baselines.}
We compared our proposed models with six baselines:
\begin{enumerate}
\itemsep0em 
  \item \textbf{L-SVM}: 
	an SVM with linear kernel.
  \item \textbf{1HNN}: 
	a single hidden layer neural network without pretraining.
  \item \textbf{DAE+1HNN}: 
	a two-layer neural network with denoising autoencoder pretraining (DAE+1HNN). 
  \item \textbf{Undo-Bias}~\cite{Khosla2012}: 
	a multi-task SVM-based algorithm for undoing dataset bias.
	Three hyper-parameters ($\lambda, C_1, C_2$) require tuning by 10FCV.
  \item \textbf{UML}~\cite{Fang2013}:
	 a structural metric learning-based algorithm that aims to learn a less biased distance metric for classification tasks.
	 The initial tuning proposal for this method was using a set of weakly-labeled data retrieved from querying class labels to search engine.
	 However, here we tuned the hyperparameters using the same strategy as others (10FCV) for a fair comparison.
  \item  \textbf{LRE-SVM}~\cite{Xu2014}:
	 a non-linear exemplar-SVMs model with a nuclear norm regularization to impose  a low-rank \emph{likelihood matrix}.
	 Four hyper-parameters ($\lambda_1, \lambda_2, C_1,C_2$) were tuned using 10FCV.
\end{enumerate}
The last three are the state-of-the-art domain generalization algorithms for object recognition.

We report the performance in terms of the classification accuracy (\%) following Xu \etal~\cite{Xu2014}.
For all algorithms that are optimized stochastically, 
we ran independent training processes using the best performing hyper-parameters in 10 times and reported the average accuracies.
Similar to the previous experiment, we do not report the standard deviations due to their small values ($\pm  0.2$).

\vspace{-1em}
\paragraph{Results on the VLCS dataset.}
We first conducted the standard training-test evaluation using L-SVM, \ie, learning the model on a training set from one domain and testing it on a test set from another domain, to check the groundtruth performance and also to identify the existence of the dataset bias.
The performance is summarized in Table~\ref{tab:vlcs_bias}.
We can see that the bias indeed exists in every domain despite the use of $\textnormal{DeCAF}_6$, the sixth layer features of the state-of-the-art deep convolutional neural network. 
The performance gap between the best cross-domain performance and the groundtruth is large, with $\geq14\%$ difference.

We then evaluated the domain generalization performance of each algorithm.
We conducted \emph{leave-one-domain-out} evaluation, which induces four cross-domain cases.
The complete recognition results are shown in Table~\ref{tab:vlcs_acc}.
In general, the dataset bias can be reduced by all algorithms after learning from multiple source domains (compare, \eg,  the minimum accuracy over the first row --V as the target-- in Table~\ref{tab:vlcs_acc} with the maximum cross-recognition accuracy over the VOC2007's column in Table~\ref{tab:vlcs_bias}).
Furthermore, Caltech-101, which is object-centric, appears to be the easiest dataset to recognize, consistent with an investigation in \cite{Torralba2011}: scene-centric datasets tend to generalize well over object-centric datasets.
Surprisingly, the performance of 1HNN has already achieved competitive accuracy compared to more complicated state-of-the-art algorithms, Undo-Bias, UML, and LRE-SVM.
Furthermore, D-MTAE outperforms other algorithms on three out of four cross-domain cases and on average, while MTAE has the second best performance on average.

\vspace{-1em}
\paragraph{Results on the Office+Caltech dataset.}
We report the experiment results on the Office+Caltech dataset.
Table~\ref{tab:office_acc} summarizes the recognition accuracies of each algorithm over four cross-domain cases.
D-MTAE+1HNN has the best performance on two out of four cross-domain cases and ranks second for the remaining cases.
On average, D-MTAE+1HNN has better performance than the prior state-of-the-art on this dataset, LRE-SVM~\cite{Xu2014}.

\vspace{-0.5em}
\section{Conclusions}
\label{sec:conclusion}
We have proposed a new approach to multi-task feature learning that reduces \emph{dataset bias} in object recognition.
The main idea is to extract features shared across domains via a training protocol that, given an image from one domain, learns to reconstruct analogs of that image for all domains.
The strategy yields two variants: the Multi-task Autoencoder (MTAE) and the Denoising MTAE which incorporates a denoising criterion.
A comprehensive suite of cross-domain object recognition evaluations shows that the  algorithms successfully learn domain-invariant features, yielding state-of-the-art performance when predicting the labels of objects from unseen target domains.

Our results suggest several directions for further study. Firstly, it is worth investigating whether stacking MTAEs improves performance. Secondly, more effective procedures for handling unbalanced samples are required, since these occur frequently in practice. Finally, a natural application of MTAEs is to streaming data such as \emph{video}, where the appearance of objects transforms in real-time.


The problem of dataset bias remains far from solved: the best model on the VLCS dataset achieved accuracies less than $70\%$ on average. A partial explanation for the poor performance compared to supervised learning is insufficient training data: the class-overlap across datasets is quite small (only 5 classes are shared across VLCS). Further progress in domain generalization requires larger datasets.

{\small
\bibliographystyle{ieee}
\bibliography{iccvbib}
}

\end{document}